\pdfoutput=1
\documentclass[11pt]{article}

\usepackage[preprint]{acl}

\usepackage{times}
\usepackage{latexsym}

\usepackage[T1]{fontenc}
\usepackage[utf8]{inputenc}

\usepackage{microtype}

\usepackage{inconsolata}

\usepackage{graphicx}

\usepackage{booktabs}
\usepackage{multirow}
\usepackage{makecell}
\usepackage{xcolor}
\usepackage{pifont}

\newcommand{\cmark}{\textcolor{green}{\ding{51}}} 
\newcommand{\xmark}{\textcolor{orange}{\ding{55}}}
\newcommand{\xxmark}{\textcolor{red}{\ding{55}}}

\usepackage{tikz}
\usetikzlibrary{shapes,arrows,positioning, bending}

\usepackage{hyphenat}      % makes hyphens in \texttt breakable
\usepackage{xspace}             % smart spacing after macros

\newcommand{\Llama}{Llama\hyp{}3.1\hyp{}8B\xspace}
\newcommand{\LlamaBig}{Llama\hyp{}3.3\hyp{}70B\xspace}
\newcommand{\PhiModel}{Phi\hyp{}4\hyp{}mini\xspace}
\newcommand{\PhiBigModel}{Phi\hyp{}4\xspace}
\newcommand{\Qwen}{Qwen2.5\hyp{}7B\xspace}
\newcommand{\QwenBig}{Qwen2.5\hyp{}14B\xspace}

\usepackage[most]{tcolorbox}
\newtcolorbox{promptbox}{
  colback=gray!5!white,
  colframe=gray!80!black,
  fonttitle=\bfseries,
  boxrule=0.5pt,
  arc=4pt,
  breakable
}

\usepackage{subcaption} 

\usepackage{pgfplots}
\pgfplotsset{compat=1.18}
\usepackage{fix-cm}  % Allows to set float-based font-size, used in PGF plots

\everymath=\expandafter{\the\everymath\displaystyle}

\title{\textit{Time} to Revisit Exact Match}

\author{Auss Abbood \\
  Language Technology Lab \\ 
  University of Cambridge \\
  \texttt{aa2613@cam.ac.uk} \\\And
  Zaiqiao Meng \\
  School of Computing Science \\ 
  University of Glasgow  \\
  \texttt{zaiqiao.meng@glasgow.ac.uk} \\\And
  Nigel Collier \\
  Language Technology Lab \\ 
  University of Cambridge \\
  \texttt{nhc30@cam.ac.uk} \\}

\begin{document}
\maketitle
\begin{abstract}
    Temporal question answering is an established method for evaluating temporal reasoning in large language models. Expected answers are often numeric (e.g., dates or durations), yet model responses are evaluated like regular text with exact match (EM), unable to distinguish small from large errors. In this investigative work, we frame temporal question answering as a numerical estimation task to assess the shortcomings of EM. We introduce \textit{TempAnswerQA}, a benchmark distilled from Test of Time and TempTabQA, where all questions require a numerical, temporal answer, allowing us to evaluate models beyond EM. We use the forecasting metrics symmetric mean absolute percentage error (sMAPE) and mean absolute scaled error (MASE). With sMAPE, we find that error size and EM are decoupled. Models with low EM still have low sMAPE (both ~20\%), and some models have high sMAPE despite high EM. Scaling errors by the deviation of the ground truth data with MASE reshuffles model rankings compared to EM, revealing gaps in models' understanding of temporal domain knowledge, especially when trained with synthetic data. Lastly, the models' most frequent error is to deviate by only $\pm1$ from the ground truth. sMAPE and MASE, unlike EM, adequately weight these errors. Our findings~\footnote{\url{https://github.com/aauss/temporal-answer-qa}} underscore the need for specialised metrics for temporal QA tasks.
\end{abstract}

\section{Introduction}

Time is an inherent part of the real world, and reasoning about it is essential for intelligent behaviour \citep{xiong_large_2024}. As such, temporal reasoning is crucial in many domains, including high-stakes areas such as logistics \citep{li_large_2023}, finance \citep{wu_bloomberggpt_2023}, and medicine \citep{blease_generative_2024}, which increases the stakes for adequate evaluation. Temporal question-answering (QA) benchmarks are a well-established method for evaluating temporal reasoning in large language models (LLMs), and the binary string-matching metric exact match (EM) is a widely used for this purpose \citep{wang_tram_2024, wei_menatqa_2023}.

\begin{figure}[t!]
	\centering
	\small
	\begin{tabular}{@{}lll@{}}
		\toprule
		\multicolumn{3}{c}{\textbf{Exact match}}                                                                                                               \\
		\midrule
		\multicolumn{3}{@{}l}{%
		\makecell[l]{\textbf{Q1}: How many hours before an anaesthesia with                                                                                    \\Halothane should you stop taking Levodopa?}
		}                                                                                                                                                      \\

		\textbf{A1}: 8                    & Model A: 8 \cmark                                                                            & Model B: 24 \xxmark \\
		\addlinespace[0.7em]
		\multicolumn{3}{@{}l@{}}{%
		\makecell[l]{\textbf{Q2}: What is the absolute time difference between Andi                                                                            \\and Lee in hours given Andi is in EST(-0500)\\and Lee is in PST(-0800)?}}\\

		\textbf{A2}: 3                    & Model A: 5 \xmark                                                                            & Model B: 3 \cmark   \\
		\addlinespace[0.7em]
		\textbf{\makecell[l]{Conclusion}} & \multicolumn{2}{@{}l@{}}{Model A and B tie on exact match rate.}                                                   \\
		\toprule
		\multicolumn{3}{c}{\textbf{Temporal difference}}                                                                                                       \\
		\midrule
		\addlinespace[0.5em]
		\textbf{\makecell[l]{Q1:}}        & \multicolumn{2}{@{}l}{\begin{tikzpicture}[scale=0.45,baseline=(current bounding box.center)]
				                                                          \draw[gray, thick, ->] (0,0) -- (12,0);
				                                                          \draw[red, very thick, -] (4,0) -- (12,0);
				                                                          \foreach \x in {0,0.5,...,12}{
						                                                          \draw (\x,2pt) -- (\x,-2pt);
					                                                          }

				                                                          \foreach \x/\lab in {4/8,12/24}{
						                                                          \node[
							                                                          anchor=south,
							                                                          font=\small,
							                                                          execute at begin node={\ifnum\lab=8\bfseries\fi}
						                                                          ] at (\x,1.5pt) {\lab};
					                                                          }
				                                                          \node[
				                                                          draw=black,
				                                                          fill=green,
				                                                          isosceles triangle,
				                                                          rotate=-90,
				                                                          inner sep=1pt,
				                                                          label=left:{\scriptsize Expected},
				                                                          ] at (4,1.3) {};
				                                                          \node[
				                                                          draw=green,
				                                                          fill=green,
				                                                          isosceles triangle,
				                                                          rotate=90,
				                                                          inner sep=1pt,
				                                                          label=above:{\scriptsize Model A},
				                                                          ] at (4,-0.7) {};
				                                                          \node[
				                                                          draw=red,
				                                                          fill=red,
				                                                          isosceles triangle,
				                                                          rotate=90,
				                                                          inner sep=1pt,
				                                                          label=above:{\scriptsize Model B}
				                                                          ] at (12,-0.7) {};
			                                                          \end{tikzpicture}     }                        \\
		\addlinespace[0.7em]
		\textbf{\makecell[l]{Q2:}}        & \multicolumn{2}{@{}l}{\begin{tikzpicture}[scale=0.45,baseline=(current bounding box.center)]
				                                                          \draw[gray, thick, ->] (0,0) -- (12,0);
				                                                          \draw[orange, very thick, -] (1.5,0) -- (2.5,0);

				                                                          \foreach \x in {0,0.5,...,12}{
						                                                          \draw (\x,2pt) -- (\x,-2pt);
					                                                          }

				                                                          \foreach \x/\lab in {1.5/3,2.5/5}{
						                                                          \node[
							                                                          anchor=south,
							                                                          font=\small,
							                                                          execute at begin node={\ifnum\lab=3\bfseries\fi}
						                                                          ] at (\x,1.5pt) {\lab};
					                                                          }

				                                                          \node[
				                                                          draw=black,
				                                                          fill=green,
				                                                          isosceles triangle,
				                                                          rotate=-90,
				                                                          inner sep=1pt,
				                                                          label=left:{\scriptsize Expected},
				                                                          ] at (1.5,1.3) {};
				                                                          \node[
				                                                          draw=green,
				                                                          fill=green,
				                                                          isosceles triangle,
				                                                          rotate=90,
				                                                          inner sep=1pt,
				                                                          label=left:{\scriptsize Model B},

				                                                          ] at (1.5,-0.5) {};
				                                                          \node[
				                                                          draw=orange,
				                                                          fill=orange!70,
				                                                          isosceles triangle,
				                                                          rotate=90,
				                                                          inner sep=1pt ,
				                                                          label=below:{\scriptsize Model A},

				                                                          ] at (2.5,-0.5) {};
			                                                          \end{tikzpicture}     }                        \\
		\addlinespace[0.7em]
		\textbf{\makecell[l]{Conclusion}} & \multicolumn{2}{@{}l@{}}{Model A has a smaller error than Model B. }                                               \\
		\bottomrule
	\end{tabular}
	\caption{\label{tab:example_metric} Exemplary performance evaluation of two models comparing exact match and temporal difference. Both models have an exact match of 50\%, but Model B has a greater temporal difference than Model A.}
\end{figure}
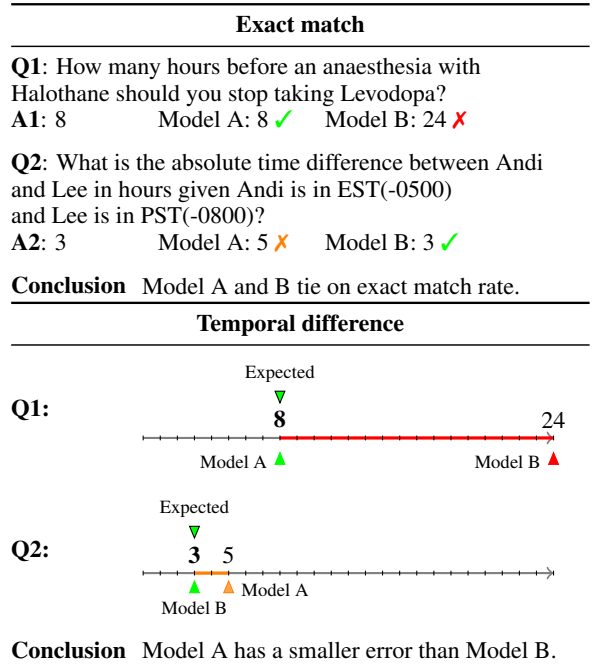

While prevalent, EM does not consider the continuous nature of time. As illustrated in Fig.~\ref{tab:example_metric}, EM considers Model A and Model B to be tied, despite Model A's error being much smaller ($\Delta$2h) than Model B's ($\Delta$8h). Although continuous alternatives exist, such as ROUGE, METEOR~\citep{gupta_temptabqa_2023} and F1 \citep{gruber_complextempqa_2024}, they collapse to binary scores when temporal answers consist solely of digits. The limitations of current metrics have been noted across domains. In medication direction systems, for example, these metrics are unable to distinguish acceptable from lethal errors in medication schedules \citep{pais_large_2024} (e.g., Q1 in Fig.~\ref{tab:example_metric}). Therefore, current benchmarks suffer from a mismatch between evaluation and deployment risk. This work aims to address this mismatch by exploring metrics more suitable for the temporal nature of the task.

Exploring continuous alternatives to EM allows us to differentiate between small and large errors. Beyond that, continuous metrics are more suitable for assessing temporal reasoning for two additional reasons. First, studies by \citet{jack_lindsey_biology_2025} and \citet{khodja_factual_2025} have shown that LLMs tend to approximate the answer to a (temporal) arithmetic task. Relying solely on EM undervalues models that approximate correct answers well. Second, answers to temporal questions can be ambiguous, such as calculating a person's age using only their birth year, where two answers with a difference of 1 year could be true \citep{khodja_factual_2025}. This ambiguity is caused by transitional times. With EM alone, we cannot distinguish relevant errors from transitional time ambiguities.

We frame temporal QA as a numerical estimation task and borrow two scale-free error metrics from forecasting to evaluate LLMs beyond EM. The first is the symmetric mean absolute percentage error (sMAPE) \citep{tofallis_better_2015}, which measures the percentage error of the model predictions. The second is the mean absolute scaled error (MASE) \citep{hyndman_another_2006}. This metric scales errors by a sensible baseline derived from the benchmark data, thus aiming to measure the models' temporal domain knowledge.

Our contributions can be summarised as follows:

\begin{enumerate}
	\item We sample QA pairs from recent temporal benchmarks composed solely of questions requiring temporal answers to explore the limitations of EM. Augmenting questions with metadata allows us to transform model responses into time-aware objects that are suitable for regression-based metrics.
	\item Our evaluation with the regression-based metric sMAPE reveals that relative errors do not increase much even for very low EM (both $\sim$20 \%). At the same time, it reveals outliers, that is, models with large relative errors despite a high EM. EM and sMAPE produce similar but not identical model rankings, making it a crucial addition to identifying robust models that make smaller errors.
	\item MASE scales errors by the deviation of the ground truth data to assess the temporal domain knowledge of the models. It yields different model rankings than EM, lowering the ranking of models trained on synthetic data. MASE reveals that models can achieve high EM and sMAPE and still make errors that exceed what we expect, given sufficient temporal domain knowledge.
	\item Finally, by treating errors numerically, we show that many model predictions are off by only $\pm1$, caused by transitional times (e.g., determining someone's age based only on their birth year). Furthermore, MASE shows that the error magnitude is not symmetric to the sign, and that errors with a positive sign are significantly larger ($>0$). Our findings underscore the need for a specialised evaluation procedure for temporal QA tasks and the inadequacy of using EM alone.

\end{enumerate}

\section{Related work}

\subsection{Temporal QA benchmarks}
Generally speaking, temporal QA aims to evaluate a model's understanding of time. Prior work often thematises the numeric nature of this task. The seminal QA benchmark TempQuestions by \citet{jia_tempquestions_2018} defined temporal questions as those that have a temporal expression (e.g. ``three weeks''), a temporal signal (e.g. ``before''), or expect a temporal answer (``When...''). The latter indicates that the expected answer needs to be a \emph{measure} of time. \citet{tan_towards_2023}, while probing temporal reasoning in LLMs, identified a connection between a lack of temporal reasoning and shortcomings in numeric reasoning. Again, this highlights the central role of numeric properties in time in temporal QA. Furthermore, temporal reasoning capabilities have often been linked to mathematical reasoning skills \cite{su_timo_2024,yuan_back_2024, fatemi_test_2025, wang_tram_2024,islakoglu_chronosense_2025}. While there is consensus on the numeric properties of time, it has not been studied in isolation.

\subsection{Evaluation challenges in temporal QA}
All benchmarks mentioned above either use token-level binary metrics or EM for evaluation. In one instance, ROUGE and METEOR were also used \citep{gupta_temptabqa_2023}.

Non-binary evaluations were conducted in some instances. \citet{tan_towards_2023} and \citet{wang_towards_2025} measured the mean absolute error for a selection of temporal arithmetic tasks. However, this measure cannot be compared across temporal resolutions (days vs. years). \citet{wallat_study_2025} proposed a date-matching metric. However, their experiments focused on event dating and robustness against differing time granularities. \citet{tan_towards_2023} also measured trend accuracy, recognising that temporal errors are directional. Since this metric is binary, it does not detect directional biases. 

Evaluations of models in an application setting are less informative because established metrics do not consider the numeric nature of time \citep{pais_large_2024}. \citet{zhang_libra_2025} mitigated this issue by using a temporal version of the F1 score that considers only temporal entities. This score is adequate for evaluating longer texts, but not in settings where answers consist only of digits, such as our own.

A review by \citet{su_temporal_2024} shows that a growing body of work in temporal QA focuses on knowledge graphs. They often aim to retrieve the correct answer from graphs. Retrieval is evaluated differently from free text, so the concerns raised in this work do not apply here.

\subsection{Transitional times}
The necessity of investigating error magnitudes has been shown before. \citet{khodja_factual_2025} found that LLMs have a significantly higher log-likelihood for answers constituting transitional times (errors of $\pm1$) than for the correct answer. They hypothesised that transitional dates are more prevalent in the models' training data since events tend to be mentioned more often around their start and end. However, the log-likelihood of answers is not available for closed-source models.

\citet{fatemi_test_2025} also observed a higher proportion of errors equal to $\pm1$ in duration questions and suspected shortcomings in the models' arithmetic precision. Despite these findings, no alternative to EM has been proposed. We, therefore, see an urgent need to investigate model errors on a continuous scale.

\section{Methods and data}

\subsection{Dataset creation}

Existing temporal QA benchmarks expect a mix of free text and temporal answers. ``Who won the Oscar for best actor in 2024?'' is a temporal question, but its answer is not. ``When was Oppenheimer released?'', on the other hand, expects a temporal answer. We classified an answer as a \emph{temporal answer} if it is a date or a duration (including age). Currently, no QA dataset expects only (numeric) temporal answers. To fill this gap, we sampled a QA dataset that expects only temporal answers, which we refer to as \textit{TempAnswerQA}. Tab.~\ref{tab:data-labeling} contains an example of the dataset.

\begin{table}
	\footnotesize
	\centering
	\renewcommand{\arraystretch}{3}
	\begin{tabular}{@{}lccc@{}}
		\toprule
		Question & Answer & \textit{\makecell{Temporal \\answer}} & \textit{\makecell{Answer\\format}}  \\
		\midrule
		\makecell[l]{How many years did Art            \\Carney work as an actor\\starting from 1939?} &  54 &  \ding{51} &  \# years  \\
		\makecell[l]{Who was the spouse                \\of Art Carney in 1970?} &  \makecell{Barbara\\Isaac} &  \ding{55} & --  \\
		\bottomrule
	\end{tabular}
	\caption{\label{tab:data-labeling} Example of labelling results for TempAnswerQA. Questions from TTQA and ToT expecting a temporal answer (date or duration) were retained. The expected answer format was added to facilitate parsing answers as numeric objects. Newly created columns are in italics.}
\end{table}

The dataset should reflect current benchmarks and, therefore, should include stand-alone questions, questions that require context, real-world questions, and synthetic ones. The latter has become increasingly relevant for combating leakage into LLMs' training data. Test of Time (ToT) and TempTabQA (TTQA) meet these requirements \footnote{ToT and TTQA have CC BY 4.0.}.

ToT \citep{fatemi_test_2025} is a synthetic QA benchmark for temporal reasoning. It consists of an arithmetic and a semantic subset. The arithmetic subset has a real-world focus and contains questions that require time-related computations. The semantic subset consists of questions related to randomly generated graphs that assess the model's understanding of temporal semantics and logic.

The enhanced version of TTQA evaluates a model's ability to answer temporal questions over semi-structured Wikipedia tables \citep{deng_enhancing_2025}. The authors split the dataset to mitigate data leakage problems into a head and tail dataset, where the latter consists of less-frequented tables.

We manually extracted questions that require a temporal answer, resulting in 1,103 QA pairs for the head subset of TTQA and 634 for the tail subset. For ToT, we extracted 1,016 QA pairs for the arithmetic subset and 681 for the semantic subset. In total, we have 3,434 QA pairs. Additionally, we annotated the required temporal unit for each question, i.e. if the answer is a date or a temporal measure in years, months, days, minutes, or seconds. We chose the higher temporal resolution if the answer contained a mix of units, for example, seconds if the answer was formatted as HH:SS. Lastly, we annotated the expected answer format to allow parsing the answer numerically as integers, \texttt{timedelta}, or \texttt{datetime} objects in Python. Tab.~\ref{tab:temp-unit-stat} lists the number of temporal answer units per dataset.

\begin{table}
	\centering
	% \small
	\begin{tabular}[t]{@{}lr@{}}
		\toprule
		\multicolumn{2}{c}{ToT} \\
		\midrule
		\makecell{Temporal      \\ unit} & Count   \\
		\midrule
		\# seconds & 411        \\
		Date       & 328        \\
		\# years   & 229        \\
		\# days    & 100        \\
		\# months  & 50         \\
		\# minutes & 38         \\
		\bottomrule
	\end{tabular}
	\hspace{1em}
	\begin{tabular}[t]{@{}lr@{}}
		\toprule
		\multicolumn{2}{c}{TTQA} \\
		\midrule
		\makecell{Temporal       \\ unit} & Count    \\
		\midrule
		\# years  & 1194         \\
		yyyy      & 305          \\
		\# days   & 94           \\
		\# months & 85           \\
		Date      & 59           \\
		\\
		\bottomrule
	\end{tabular}
	\caption{\label{tab:temp-unit-stat}The number of questions per temporal unit of the answer. Answers can be either a duration measured as a number of <temporal unit>, a full date or a date with only the year (yyyy).}
\end{table}

\subsection{Regression-based metrics for temporal QA}

Metrics used to evaluate QA benchmarks are designed for text and, therefore, do not capture the size and direction of the error for temporal answers. Specifically, minor errors due to transitional times are indistinguishable from significant errors. Temp\-Ans\-wer\-QA's expected answers all have numerical representation, which allows us to use regression-based metrics for evaluation.

There were a few considerations we made before selecting metrics. We needed to select (1) an aggregation technique that avoids errors of different signs cancelling each other out, (2) decide whether we want to weight errors, (3) how to summarise errors, and lastly, ensure that (4) errors will be comparable across different units, e.g., years and seconds.

We selected metrics using absolute errors to avoid the cancellation of errors of different signs. We decided against weighting errors by squaring or taking their logarithm, as this impedes interpretation, and we lack justification. Errors can be summarised using the mean or median. However, the median resulted in many scores of 0s or 100s for EM and sMAPE. Therefore, we picked the mean. Lastly, we needed to select a scale-free metric to compare errors across units (e.g. relative errors).

sMAPE is scale-free and uses absolute errors. It is bounded between 0 and 100 and exhibits higher symmetry between negative and positive errors than its precursor, the mean absolute percentage error, although a bias against under-predictions remains. sMAPE cannot be easily compared between experiments as its denominator contains model predictions and expected values. It is defined as:

$$sMAPE = \frac{100\%}{n} \sum_{i=1}^{n} \frac{|\hat{y}_i - y_i|}{|\hat{y}_i| + |y_i| },$$

where $n$ is the number of QA pairs, $y$ is the expected temporal answer, and $\hat{y}$ is the predicted temporal answer. If an answer is not parsable, sMAPE is defined as 100\%, and if the numerator and denominator are 0, we define it as 0\%.

A subset of answers are dates whose percentage error is not defined (Tab.~\ref{tab:temp-unit-stat}). Therefore, we also consider MASE. It fulfils our requirements and is defined for dates. MASE measures the absolute errors scaled by the mean absolute deviation of the dataset. It has no upper bound like sMAPE. MASE is also considered superior to most forecasting metrics and is used in the well-known Makridakis forecasting challenge \citep{makridakis_m5_2022}. We use an adaptation for non-timeseries data \citep{hyndman_forecasting_2014}. It is defined as:

$$MASE = \frac{1}{n} \sum_{i=1}^{n} \frac{|\hat{y}_i - y_i|}{|y_i - \bar{Y}_u|},$$

where $\bar{Y}$ is the average of the expected values. Instead of using all data to calculate $\bar{Y}$, we use a temporal unit-specific $\bar{Y}_u$. Some of the answers have a bimodal distribution. The answers with the temporal unit years in ToT have a peak for the answers $<100$ (e.g., age) and a peak around 2000 (calendar year). The mean is not representative in this case. To resolve this issue, we perform clustering with a setting that allowed our model to also return one cluster (unimodally distributed). The results and model settings are in the Appendix~\ref{sec:clustering}.

The motivation for using MASE is that the dataset’s answers are not uniformly distributed. With sufficient domain knowledge, we can often make reasonable estimates for the answer --- for instance, someone’s age is unlikely to exceed 100. Even without annotations, MASE captures such expectations from the data. However, when no plausible range exists (for example, predicting when to sell a stock), sMAPE is more interpretable.  

Another class of metrics measures semantic similarity. BERTScore \citep{zhang_bertscore_2020} is a widespread implementation of such a metric. However, it cannot distinguish between small and large differences between integers (Appendix \ref{sec:bertscore}), so we did not consider it.

\subsection{Models and prompts}

Similar to previous work, we used a selection of open-source models for our experiments, namely \PhiModel, \PhiBigModel \citep{abdin_phi-4_2024}, \Llama, \LlamaBig \citep{grattafiori_llama_2024}, \Qwen, and \QwenBig \citep{qwen_qwen25_2025}. The model settings are in the Appendix~\ref{sec:model_settings}. Since evaluation relied on parsing answers into time-aware objects, we selected instruction-tuned models for better instruction-following capabilities. We considered using Timo, a temporal Llama 2 model by \citeauthor{su_timo_2024}, but its context window was too small for some questions.

TTQA and ToT come with their own (user) prompts, which we adopted to make use of chat templates. Our selection of small models had difficulties following instructions otherwise. We moved the formatting instructions to the system prompt. These were especially important for ToT, where answers needed to be JSONs. Examples were presented as turns between the assistant and user in the case of few-shot prompting. Both adjustments improved instruction following. Furthermore, models produced valid JSONs more often when ending the prompt with an assistant turn, appending the beginning of the required JSON and removing generation prompts (see Appendices \ref{sec:ttqa_prompts} and \ref{sec:tot_prompts} for prompts and \ref{sec:model_settings} for experiments justifying chat templates and different generation strategies).

\section{Experiments and results}

We conducted our experiments based on these six selected models of different sizes, with and without few-shot prompting, on TempAnswerQA. Its questions expect temporal answers that can be assessed in a regression-like fashion. Our experiments aim to answer the following questions:

\begin{description}
	\item[RQ1:] Is the binary metric EM enough to evaluate LLMs on temporal QA benchmarks, expecting temporal answers?
	\item[RQ2:] Can regression-based metrics help improve our understanding of LLMs' performance on QA tasks expecting a temporal answer?
	\item[RQ3:] What advantages do we have in using regression-based metrics compared to EM?
\end{description}

\subsection{Exact match does not capture error magnitudes}

EM does not differentiate between small and large errors, that is, their error magnitudes. Wrong predictions ($\text{EM}=0$) can have vastly different values for sMAPE. For example, two models with EM of 80\% could have an sMAPE of 1\% and 20\%, respectively. The lower the EM rate, the wider the range of values that sMAPE can assume. Appendix~\ref{sec:smape_intuition} contains an illustration of this relationship.

\begin{figure*}[t!]
	\input{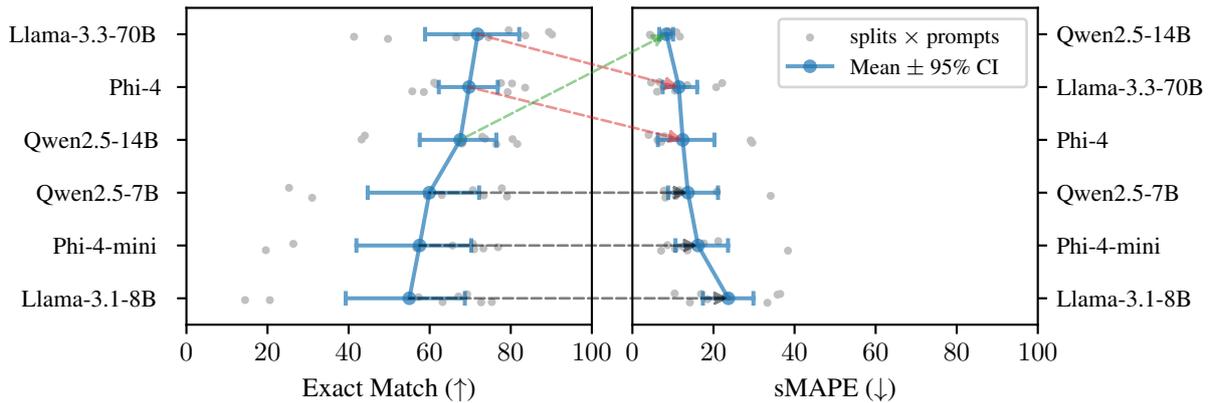}
	\caption {\label{fig:smape_em_results}Model ranking by sMAPE and EM. Blue dots represent the mean score, and bars around them the 95\% confidence interval. Grey dots are individual runs with and without few-shot prompting on all splits of ToT and TTQA. Arrows indicate a rank change from EM to sMAPE. It is green if it improves, red if it decreases, and black if it stays the same.}
\end{figure*}

Model predictions on the TempAnswerQA dataset evaluated by EM and sMAPE are shown in Fig.~\ref{fig:smape_em_results}. According to EM, \LlamaBig is the best model. \PhiBigModel and then \QwenBig closely follow it. Smaller models follow thereafter. The range of EM is wide with values as low as 20\% for \Llama, \Qwen, and \PhiModel. sMAPE values, on the other hand, span a shorter range between models and data splits (up to 40\%) than in the EM dimension (15-80\%). sMAPE changes the model ranking, placing \QwenBig in the first place. It is also the model with the narrowest 95\% confidence interval. All models, except \QwenBig, have outliers hovering around 40\%. For example, \LlamaBig fails severely in answering how many days Ingenuity took to reach Mars. Due to an arithmetic mistake, it answers 0.057 days. The expected answer is 418 days.

\QwenBig, which has improved mathematical capabilities and improved understanding of structured data, overtakes \LlamaBig when evaluated with sMAPE. The findings also show that larger models perform better, equivalent to the EM results. If errors produce non-linear costs and low errors are more desired than a high EM, \QwenBig should be preferred over \LlamaBig. The results in tabular form, including baselines predicting the mean and median, are in Appendix~\ref{sec:appendix-results}. This appendix also contains an investigation of the relationship between correct model responses and arithmetic mistakes in the chain-of-thought (CoT) reasoning traces for the ToT dataset.

\subsection{Tolerable error magnitudes depend on the task difficulty}

MASE was introduced as a metric superior to other regression-based metrics and is the gold standard in forecasting. Unlike sMAPE, it can also be applied to dates. Its main property is that it scales the prediction errors by the difficulty of the problem, which is relevant because answers in the TempAnswerQA are not arbitrarily distributed and benefit from temporal domain knowledge. For example, a subset of questions is related to the time zone. The maximum time difference between time zones is 26 hours. Models with this knowledge should not produce errors larger than that. Without human annotation for acceptable error ranges, MASE can extract them from the data instead.

\begin{figure*}[t!]
	\input{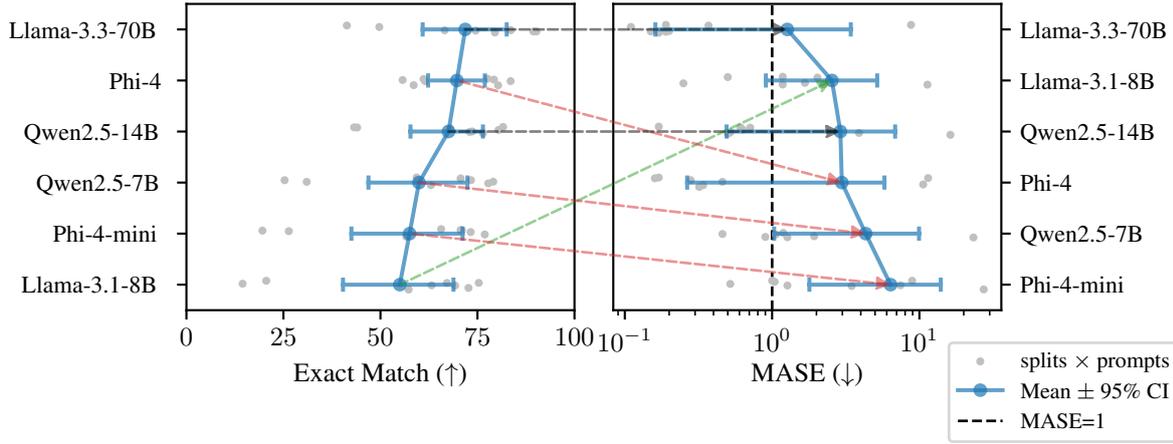}
	\caption {\label{fig:mase_em_results}Model ranking by MASE and EM. Blue dots represent the mean score, and bars around them the 95\% confidence interval. Grey dots are individual runs with and without few-shot prompting on all splits of ToT and TTQA. Arrows indicate a rank change from EM to MASE. It is green if it improves, red if it decreases, and black if it stays the same.}
\end{figure*}

Figure~\ref{fig:mase_em_results} shows model performance on TempAnswerQA using EM and MASE. All models have a MASE above 1, indicating their mean absolute error exceeds the dataset’s mean absolute deviation (stratified by temporal unit and data split). While EM and sMAPE tend to favour larger models—and sMAPE highlighted \QwenBig’s strength from training on maths and structured data, MASE tells a different story. \Llama jumps from last place to second. This shift illustrates why error scaling matters: sMAPE alone does not show whether a mistake is significant given plausibility ranges for answers. For example, when asked in which year racing driver Jenson Button won his first championship, \Qwen answers 2018 instead of 2009. That nine-year gap yields a scaled error of 5.12, which is given the brevity of athletic careers.

\Qwen's and \Llama's MASE scores differ significantly, despite similar architectures and parameter sizes, leaving the difference in training data as an explanation for performance gaps. Interestingly, Llama models are the only ones not trained on synthetic data. We thus suspect Qwen's and Phi's synthetic training regimes distorts the models' temporal domain knowledge, comparable to catastrophic forgetting. The tabular results, including baselines that predict the mean and median, are presented in Appendix~\ref{sec:appendix-results}. Exemplary model responses are provided in Appendix~\ref{sec:model-responses}.

\subsection{Scaled errors produce different rankings, percentage errors do not}

\begin{figure}
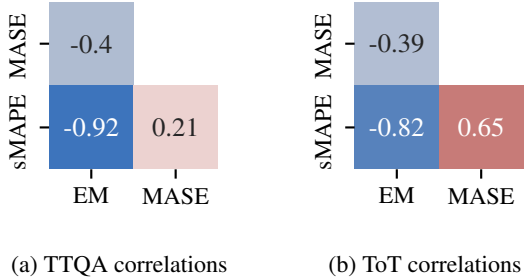

	\centering
	\begin{subfigure}[b]{0.23\textwidth}
		\centering
		\input{figures/ttqa_rank_corr.pgf}
		\caption{TTQA correlations}
		\label{fig:ttqa_rank_corr}
	\end{subfigure}
	\hfill
	\begin{subfigure}[b]{0.23\textwidth}
		\centering
		\input{figures/tot_rank_corr.pgf}
		\caption{ToT correlations}
		\label{fig:tot_rank_corr}
	\end{subfigure}
	\caption{Spearman rank correlation between metrics on all experiments (models $\times$ prompts) per dataset.}
\end{figure}

EM is a gold standard metric for evaluating LLMs on QA benchmarks. Therefore, it is necessary to compare sMAPE and MASE with EM. We used Spearman's rank correlation coefficient to compare model rankings across metrics, and the results are shown in Fig.~\ref{fig:ttqa_rank_corr} and Fig.~\ref{fig:tot_rank_corr}.

EM has a high rank correlation with sMAPE for both datasets, ToT (-0.82) and TTQA (-0.92). It is negative because a higher EM is better, while a lower sMAPE is better. The correlation is much lower between MASE and EM, with values around 0.4 for both datasets.

Considering the high agreement in the ranking between both metrics, but knowing that sMAPE is more affected by outliers by definition, which is also observable in Fig.~\ref{fig:smape_em_results}, we find that sMAPE is a crucial addition to EM for model evaluation if error magnitude matters. Since it does not produce significantly different model ranks, interpreting EM and sMAPE in tandem is easier.

MASE produces different model ranks, which is unsurprising since, unlike sMAPE, the same error magnitude scales differently depending on the task. MASE is, therefore, stricter if data deviation is low. Scaling errors for time zone or age-related questions are examples of this. Datasets are most likely designed to span reasonable time periods. If not, clustering should help make MASE more reliable. However, further verification, ideally by humans, is required.

\subsection{Transitional times and error directions}

Casting answers into time-aware objects allows us to investigate raw errors, helping us identify off-by-one errors ($\pm1$) due to transitional times and the direction of the error. Transitional times most often involve questions asking for durations. To investigate this relationship, we measure the frequency of these off-by-one errors ($|e|=1$) and determine whether they occur more often in duration questions.

Indeed, our analysis reveals that off-by-one errors ($|e|=1$) are the most frequent in both datasets (Tab.~\ref{tab:abs_wrongs}). For ToT, the share of these errors is 11.62\%. For TTQA, it is 49.49\%. This result is significant because the number of possible errors is infinite.

\begin{table}
	\centering
	\small
	\begin{tabular}{@{}lrr@{}}
		\toprule
		\multicolumn{3}{c}{ToT}         \\
		\midrule
		$|e|$ & Count & \makecell{Share \\ (\%)} \\
		\midrule
		1     & 1002  & 11.62           \\
		2     & 446   & 5.17            \\
		4     & 344   & 3.99            \\
		3     & 258   & 2.99            \\
		5     & 208   & 2.41            \\
		\bottomrule
	\end{tabular}
	\hspace{2em}
	\begin{tabular}{@{}lrr@{}}
		\toprule
		\multicolumn{3}{c}{TTQA}        \\
		\midrule
		$|e|$ & Count & \makecell{Share \\ (\%)} \\
		\midrule
		1     & 1853  & 49.49           \\
		2     & 250   & 6.68            \\
		3     & 159   & 4.25            \\
		4     & 128   & 3.42            \\
		6     & 117   & 3.12            \\
		\bottomrule
	\end{tabular}
	\caption{\label{tab:abs_wrongs}Five most frequent absolute errors per dataset over all experiments (models $\times$ prompts) with number of occurrences and relative share in percent. Note that models performed better on TTQA, explaining the high share of errors equal to $|e|=1$.}
\end{table}

Next, we verify whether $|e|=1$ errors occur more often for duration-related questions. We divide the dataset by question type as defined by the authors of ToT, and by answer format for TTQA. Tab.~\ref{tab:tot_question_type_share} shows that the types of questions are evenly distributed within ToT. The share of question types where $|e|=1$ is vastly different. RelationDuration and Duration questions tremendously increase their share. The share for Trick questions doubles. The Trick setup confuses LLMs about whether to exclude or include either the start and end dates for a duration calculation.

\begin{table}[h]
	\centering
	\small
	\begin{tabular}{@{}lrr@{}}
		\toprule
		\multirow{2}{*}{\makecell{Question type}} & \multicolumn{2}{c}{Share (\%)}                   \\
		\cmidrule(lr){2-3}
		                                          & all data                       & where $|e| = 1$ \\
		\midrule
		MultiOP                                   & 20.57                          & 4.99            \\
		EventAtWhatTime                           & 20.15                          & 4.59            \\
		RelationDuration                          & 19.98                          & 32.14           \\
		AddSubtract                               & 14.73                          & 16.57           \\
		Duration                                  & 11.79                          & 18.96           \\
		Trick                                     & 6.89                           & 22.36           \\
		Timezone                                  & 5.89                           & 0.40            \\
		\bottomrule
	\end{tabular}
	\caption{Share of question types in ToT dataset compared by share of question types where prediction error is 1 ($|e|=1$) over all experiments (models $\times$ prompts).}
	\label{tab:tot_question_type_share}
\end{table}

Due to a lack of question-type labels in TTQA, we use the expected answer format instead. Tab.~\ref{tab:ttqa_answer_format_share} compares the share of questions by answer format for all data and when the errors are equal to $|e|=1$. The TTQA dataset contains many more duration-related questions than ToT. Therefore, the increase in share is not as prominent as in ToT, but it is striking that all non-duration answers have a significantly smaller share among the questions where the error is $|e|=1$.

\begin{table}
	\centering
	\small
	\begin{tabular}{@{}lrr@{}}
		\toprule
		\multirow{2}{*}{\makecell{Answer format}} & \multicolumn{2}{c}{Share (\%)}                   \\
		\cmidrule(lr){2-3}
		                                          & all data                       & where $|e| = 1$ \\
		\midrule
		\# years                                  & 68.74                          & 81.27           \\
		yyyy                                      & 17.56                          & 5.56            \\
		\# days                                   & 5.41                           & 7.34            \\
		\# months                                 & 4.89                           & 5.40            \\
		\%B \%d, \%Y                              & 3.40                           & 0.43            \\
		\bottomrule
	\end{tabular}
	\caption{Share of answer formats in TTQA dataset compared by share of answer formats where prediction error is 1 ($|e|=1$) over all experiments (models $\times$ prompts).}
	\label{tab:ttqa_answer_format_share}
\end{table}

Finally, we investigate whether model errors have a directional bias. In Tab.~\ref{tab:error-dir}, we see that sMAPE is similar for positive and negative errors. This is not the case for MASE. Positive errors produce much higher MASE. This difference is pronounced for the TTQA dataset. In the ToT dataset, the difference in the standard deviation is more noticeable. This insight is relevant to applications where the cost of errors is not symmetric with respect to direction.

\begin{table}
	\centering
	\small
	\begin{tabular}{@{}llcc@{}}
		\toprule
		\makecell{Dataset}       & \makecell{Error} & \makecell{sMAPE                  \\($\pm$std)} & \makecell{MASE \\($\pm$std)}  \\
		\midrule
		\multirow[t]{2}{*}{ToT}  & neg.             & 24.73 (31.21)   & 1.40 (7.09)    \\
		                         & pos.             & 21.60 (29.26)   & 3.98 (40.82)   \\
		\multirow[t]{2}{*}{TTQA} & neg.             & 22.83 (30.42)   & 0.55 (1.09)    \\
		                         & pos.             & 29.32 (31.72)   & 48.09 (334.80) \\
		\bottomrule
	\end{tabular}
	\caption{sMAPE and MASE including their standard deviation where error is either strictly positive or negative per dataset.}
	\label{tab:error-dir}
\end{table}

\section{Conclusion}

In this work, we release TempAnswerQA, a distilled benchmark focused on the continuous nature of time. With it, we show that EM systematically ignores error magnitude and direction, which are both critical to temporal reasoning.

To this end, we use sMAPE and MASE, two regression-based metrics that capture properties in the prediction errors of the models that EM does not. sMAPE is relatively low, even if EM is low. EM underestimates the models' understanding of the correct answer. \QwenBig, which was trained on structured data and mathematical reasoning, ranks first according to sMAPE, overtaking \LlamaBig, the best model according to EM. Both Llama models perform the best according to MASE. They are the only models not trained on synthetic data, suggesting that their temporal domain knowledge is higher and synthetic data distorts this knowledge.

Answers to duration-related questions can be ambiguous due to transitional times, leading to two answers being correct with a difference of just 1. This leads to an inflation of errors equal to $\pm1$. sMAPE and MASE are continuous metrics and thus provide a more balanced evaluation than EM.

Lastly, we show that MASE and sMAPE are valuable additions to EM. Although sMAPE yields rankings similar to EM, its sensitivity to error sizes gives it an edge when large deviations need to be penalised. MASE ranks models significantly differently. It attempts to scale errors by plausible ranges of correct answers and thus tries to probe models' temporal domain knowledge rather than their relative errors as with sMAPE. Without human-annotated data, MASE is a viable alternative to measure prior temporal knowledge.

\section{Outlook}

While we have shown critical gaps in EM for temporal reasoning evaluation and offered regression-based alternatives, more work is required to verify their benefits. Verification can be achieved either through a separate dataset or by human preference. Specifically, other approaches for scaling errors for MASE should be considered. Future work should also consider LLM-as-a-judge to overcome the reported limitations. Small sMAPE suggests that models have a good understanding of the problem, but struggle with precise arithmetic. Tool calling is an interesting next step in assessing whether the low performance is due to arithmetic miscalculations rather than insufficient temporal reasoning capabilities.

\section{Limitations}

There were answers in both datasets that we excluded, although they were temporal because they are not trivially evaluable (115 in total). These included date and time ranges, multiple answers, and frequencies such as ``every first Monday of the month''. The latter is related to absolute times and dates, which have a bounded error. For example, if we ask in which month Christmas is celebrated, the maximum error is 11 months. In contrast, errors for other answers in the dataset do not have an upper bound.

Neither regression-based metric is applicable to all model responses. Either because the answer is not parsable in the case of MASE, or because the answer is a date or a time (sMAPE). This shortcoming needs to be addressed.

MASE was scaled by each subset of both datasets and the expected temporal unit of the answer. Although this approach makes the reasonable assumption that the authors of the paper produced problems that are similar within a subset and that the expected temporal unit sufficiently captures similar kinds of problems, this may not always hold. Clustering could unravel such questions into more representative clusters. However, this approach does not hold up to a hand-crafted dataset, where the mean absolute deviation is neatly justified for each question.

\section*{Acknowledgement}
This work was supported by the Economic and Social Research Council [ES/Y001788/1].

\bibliography{acl_latex}

\clearpage

\appendix

\section{sMAPE intuition}
\label{sec:smape_intuition}
\begin{figure}
	\centering
	\input{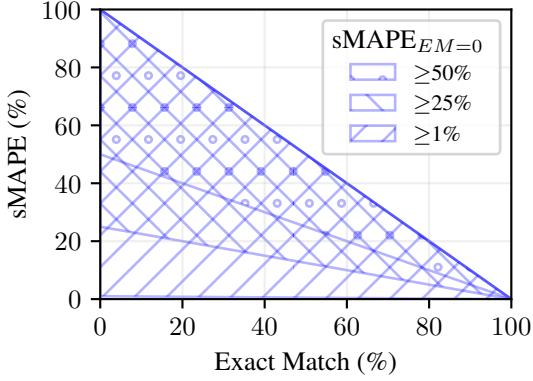}
	\caption{\label{fig:motivate_smape}EM does not capture error magnitude. The possible variance in error magnitude (measured by sMAPE) is higher the lower the EM is.}
\end{figure}

If a model's answer is wrong, the answer's error can range from a tiny relative error up to an infinitely large one. The lower the EM, the higher the sMAPE can be in a model. In other words, models with the same EM can spread more widely, and the lower the EM is, the better sMAPE is at discriminating between model performances. Fig.~\ref{fig:motivate_smape} illustrates this. If we assume that for all wrong predictions that a model makes, the minimum error measured by sMAPE is 1\%, 25\%, and 50\%, the figure shows the range of values that a model can still score with respect to sMAPE for all possible values of EM.

\section{Model settings}
\label{sec:model_settings}

All our models were accessed via Hugging Face using the \texttt{transformers} Python library at version 4.49.0 \citep{wolf_transformers_2020}. We used the default settings for each model in our experiments. For text generation, we used the settings in Tab.~\ref{tab:model-settings}. We used a mix of GPUs to run our experiments, including GeForce 3090s and 4090s, and two A100s in parallel to run inference for \LlamaBig. GPU hours required to run inference on \LlamaBig required approximately 24 hours. Experiments with smaller models took 1 to 3 hours per run. At least as many GPU hours across GPUs were used to run small experiments or to test code.

\begin{table}
	\centering
	\begin{tabular}{@{}ccc@{}}
		\toprule
		Dataset    & Max. new token & \makecell{End of \\ sequence tokens} \\
		\midrule
		ToT  & 512            & No               \\
		TTQA & 512            & Yes              \\
		\bottomrule
	\end{tabular}
	\caption{Generation settings.}
	\label{tab:model-settings}
\end{table}

The evaluation of ToT depends on the models that produce parsable JSONs. Therefore, we experimented with setting either \texttt{add\_ge\-ne\-ra\-tion\_prompt} or \texttt{con\-tin\-ue\_final\_mes\-sage} to true in Hugging Face. The first appends an assistant token to our messages, if available, indicating an answer. The latter does not do this, prompting the models to continue their messages. The resulting prompts are presented in the Appendix~\ref{sec:tot_prompts}. To test when JSON formatting was more successful, we randomly sampled 50 questions from the semantic split of ToT and compared the number of correctly parsed JSONs. The results are in Tab.~\ref{tab:last-token}. Setting \texttt{con\-tin\-ue\_final\_mes\-sage}  produced fewer parsing errors (3 over three models) than \texttt{add\_ge\-ne\-ra\-tion\_prompt} (17 over three models).

\begin{table}
	\centering
	\begin{tabular}{@{}lcc@{}}
		\toprule
		\multirow{2}{*}{Model} & \multicolumn{2}{c}{\# Parsing errors}        \\
		\cmidrule(lr){2-3}
		                       & \makecell{add generation                     \\ prompt}  & \makecell{continue \\ final message} \\
		\midrule
		Llama-3.1-8B           & 0/50                                  & 2/50 \\
		Qwen2.5-7B             & 4/50                                  & 0/50 \\
		Phi-4-mini             & 13/50                                 & 1/50 \\
		\bottomrule
	\end{tabular}
	\caption{Number of parsable JSONs per model for different generation strategies tested on 50 randomly selected questions from the semantic split of ToT.}
	\label{tab:last-token}
\end{table}

The evaluation of TTQA also depended on the correct format of the output. Specifically, models needed to place their answer after the string ``Final Answer:''. We observed a low rate of correct formatting and thus experimented with transferring prompts into a chat template. The correct output formatting was compared between the original and the prompts translated into chat templates. We tested the models' instruction following on the head split of the TTQA dataset. The results are shown in Tab.~\ref{tab:chat-template}. Qwen and Phi improved their instruction following, with Qwen almost doubling it from 44.52 to 99.56\%. Llama experiences a slight decrease in performance when using chat templates, from 81.98\% to 74.40\%.

\begin{table}
	\centering
	\begin{tabular}{@{}lcc@{}}
		\toprule
		\multirow{2}{*}{Model} & \multicolumn{2}{c}{Correct output format (\%)}                  \\
		\cmidrule(lr){2-3}
		                       & Original prompt                                & \makecell{With \\ chat template} \\
		\midrule
		Llama-3.1-8B           & 81.98                                          & 74.40          \\
		Qwen2.5-7B             & 44.52                                          & 99.56          \\
		Phi-4-mini             & 94.76                                          & 99.38          \\
		\bottomrule
	\end{tabular}
	\caption{Number of answers containing the expected string ``Final Answer:'' in their response for each model on the head split of the TTQA dataset. The percentages were calculated based on slightly varying numbers of questions as experiments were conducted at different steps in the labelling process.}
	\label{tab:chat-template}
\end{table}

\section{TTQA prompts}
\label{sec:ttqa_prompts}

Below, we list the TTQA prompts used in this work. We compare the prompts originally used by \citet{deng_enhancing_2025} and our adaptation, which utilises chat templates. For brevity, we replaced some turns in the few-shot example with ``\dots''. Furthermore, we did not use the original questions, tables, or answers below but replaced them with placeholders enclosed by ``<>''.

\subsection{TTQA zero-shot prompt}

\begin{promptbox}
	\textbf{User prompt:} Given an entity-centric table and corresponding question, answer the question by providing step-by-step reasoning and then clearly and concisely stating the final answer using "Final Answer:".

	Each table-question pair is presented as a table (identified by "Table:") followed by a question (identified by "Q:"). Tables are presented in a linear format, with columns separated by tabs, rows separated by newlines, and subsections separated by double newlines. If necessary, assume the current date is December, 2022.
	\\
	\\
	========================
	\\
	Table:
	\\
	\\
	<TABLE>
	\\
	\\
	<QUESTION>
	\\
	\\
	A: Let’s think step by step.

	\textbf{Assistant:}
\end{promptbox}

\subsection{TTQA zero-shot prompt as chat template}

\begin{promptbox}
	\textbf{System prompt:} Given an entity-centric table and corresponding question, answer the question by providing step-by-step reasoning and then clearly and concisely stating the final answer using "Final Answer:".

	Each table-question pair is presented as a table (identified by "Table:") followed by a question (identified by "Q:"). Tables are presented in a linear format, with columns separated by tabs, rows separated by newlines, and subsections separated by double newlines. If necessary, assume the current date is December, 2022.

	\textbf{User prompt:}\\
	Table:
	\\
	\\
	<TABLE>
	\\
	\\
	<QUESTION>
	\\
	\\
	A: Let’s think step by step.

	\textbf{Assistant:}
\end{promptbox}

\subsection{TTQA few-shot prompt}

\begin{promptbox}
	\textbf{User prompt:} Given an entity-centric table and corresponding question, answer the question by providing step-by-step reasoning and then clearly and concisely stating the final answer using "Final Answer:".

	Each table-question pair is presented as a table (identified by "Table:") followed by a question (identified by "Q:"). Tables are presented in a linear format, with columns separated by tabs, rows separated by newlines, and subsections separated by double newlines. If necessary, assume the current date is December, 2022.\\
	\\
	Here is an example that follows these instructions. Answer the provided questions in a similar format:\\
	\\
	========================
	\\
	Table:
	\\
	\\
	<TABLE, SHOT 1>
	\\
	\\
	Q: <QUESTION, SHOT 1>
	\\
	\\
	A: <ANSWER, SHOT 1>
	\\
	\\
	========================
	\\
	...\\
	<TABLE, SHOT 3>
	\\
	\\
	Q: <QUESTION, SHOT 3>
	\\
	\\
	A: <QUESTION, SHOT 3>
	\\
	\\
	========================
	\\
	\\
	Table:
	\\
	\\
	<TABLE>
	\\
	\\
	<QUESTION>
	\\
	\\
	A: Let’s think step by step.

	\textbf{Assistant:}
\end{promptbox}

\subsection{TTQA few-shot prompt as chat template}
\label{sec:ttqa_few_shot_chat_template}

\begin{promptbox}
	\textbf{System prompt:} Given an entity-centric table and corresponding question, answer the question by providing step-by-step reasoning and then clearly and concisely stating the final answer using "Final Answer:".

	Each table-question pair is presented as a table (identified by "Table:") followed by a question (identified by "Q:"). Tables are presented in a linear format, with columns separated by tabs, rows separated by newlines, and subsections separated by double newlines. If necessary, assume the current date is December, 2022.\\
	\\
	Here is an example that follows these instructions. Answer the provided questions in a similar format:\\
	\textbf{User prompt}\\
	Table:
	\\
	\\
	<TABLE, SHOT 1>
	\\
	\\
	Q: <QUESTION, SHOT 1>
	\\
	\\
	A:\\
	\textbf{Assistant prompt:} <ANSWER, SHOT 1>
	\\
	...\\
	\textbf{User prompt:}\\
	Table:\\
	<TABLE, SHOT 3>
	\\
	\\
	Q: <QUESTION, SHOT 3>
	\\
	\\
	A:\\
	\textbf{Assistant prompt:} <ANSWER, SHOT 3>
	\\
	\textbf{User prompt:}\\
	Table:
	\\
	\\
	<TABLE>
	\\
	\\
	<QUESTION>
	\\
	\\
	A: Let’s think step by step.

	\textbf{Assistant:}
\end{promptbox}

\section{ToT prompts}
\label{sec:tot_prompts}
Below, we list the ToT prompts used in this work. We compare the prompts originally used by \citet{fatemi_test_2025} with our adaptation, which utilises chat templates.

A few-shot version of the prompts was constructed by modifying existing questions. The chat template was filled as in \ref{sec:ttqa_few_shot_chat_template}, where examples were presented as turns between the user and the assistant. In the case of the semantic subset, the graph information was included in the system prompt. The generation prompt was removed, and the assistant prompt was pre-filled.

\subsection{ToT zero-shot prompt}

\begin{promptbox}
	\textbf{User prompt:} Natalie and Chris were born on 2004-Feb-18 and 2004-Dec-30 respectively. When Chris was 991 days old, how old was Natalie in days? Return your answer as a JSON like: JSON = \{""explanation"": <your step by step solution>, ""answer"": <num\_days>\}

	\textbf{Assistant:}
\end{promptbox}

\subsection{ToT zero-shot prompt as chat template}

\begin{promptbox}
	\textbf{System prompt:} Return your answer as a JSON like: JSON = \{"explanation": <your step by step solution>, "answer": <num\_days>\}

	\textbf{User prompt:}  Natalie and Chris were born on 2004-Feb-18 and 2004-Dec-30 respectively. When Chris was 991 days old, how old was Natalie in days?

	\textbf{Assistant:} JSON = \{"explanation":
\end{promptbox}

 \section{Cluster results}
\label{sec:clustering}

\begin{figure*}
	\centering
	\input{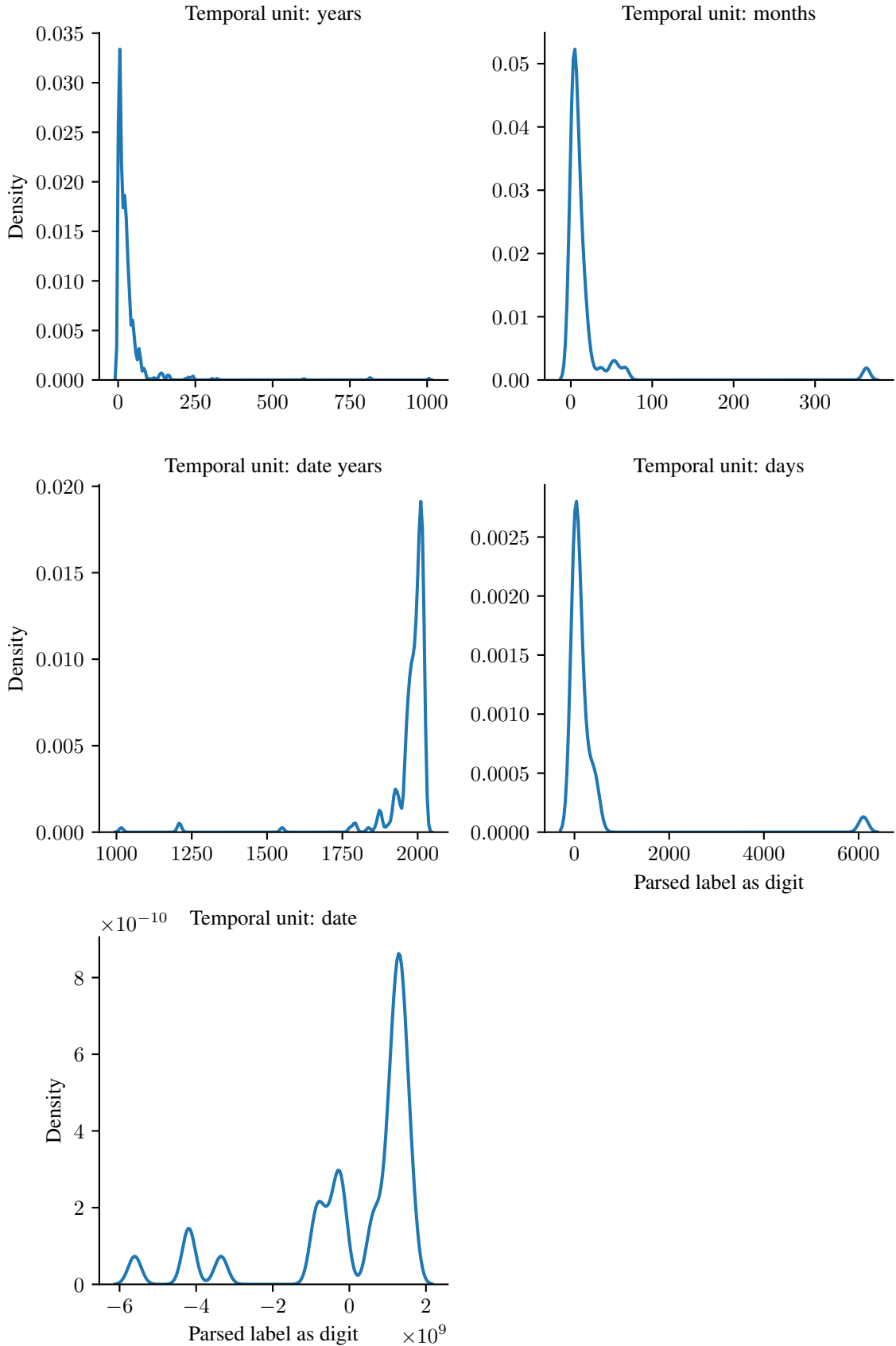}
	\caption {\label{fig:answer_kde_per_temporal_unit_ttqa_head}Distribution of the expected answers by temporal unit of the answer for the head split of the TTQA dataset. Answers were transformed into numeric form. In the case of dates, they were converted into timestamps.}
\end{figure*}

\begin{figure*}
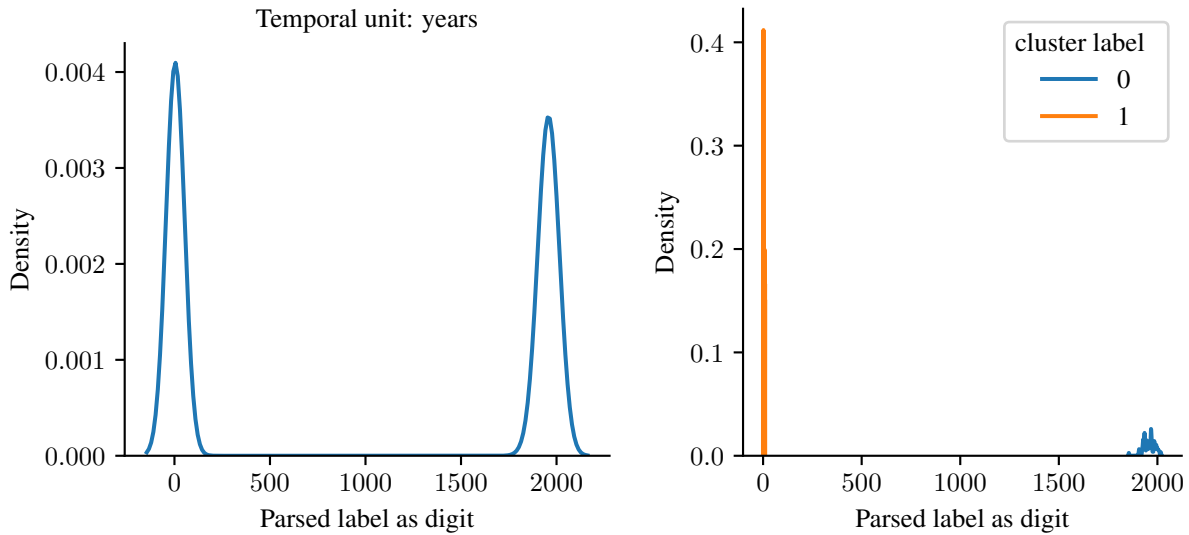

	\input{figures/answer_kde_per_temporal_unit_tot_semantic_None.pgf}
	\input{figures/answer_kde_per_temporal_unit_tot_semantic_cluster_label.pgf}
	\caption {\label{fig:answer_kde_per_temporal_unit_tot_semantic}Distribution of the expected answers by temporal unit of the answer for the semantic split of the ToT dataset. On the left is the raw distribution, and on the right is the distribution after clustering. Answers were transformed into numeric form. In the case of dates, they were converted into timestamps.}
\end{figure*}

MASE required the mean answer per temporal unit of the answer and the split of each dataset. Clustering did not affect the TTQA data. ToT, however, exhibited some bimodality, which was identified by the clustering algorithm. The distribution of the answers per split and temporal unit for TTQA is shown in Fig.~\ref{fig:answer_kde_per_temporal_unit_ttqa_head} and Fig.\ref{fig:answer_kde_per_temporal_unit_ttqa_tail}. ToT's answer distribution for the arithmetic split before and after clustering can be found in Fig.~\ref{fig:answer_kde_per_temporal_unit_tot_arithmetic} and Fig.~\ref{fig:answer_kde_per_temporal_unit_tot_arithmetic_clustered}, respectively, and in Fig.~\ref{fig:answer_kde_per_temporal_unit_tot_semantic} for the semantic split.

Clustering was performed using scikit-learn's HDBSCAN (hierarchical density-based spatial clustering of applications with noise) model. The minimum cluster size was set to 30\% to avoid too small clusters. The model was allowed to produce single clusters. All other settings were set to their default values. We used version 1.6.1 of \texttt{scikit-learn} \citep{pedregosa_scikit-learn_2011}.

\begin{figure*}
	\input{figures/answer_kde_per_temporal_unit_ttqa_tail_None.pgf}
	\caption {\label{fig:answer_kde_per_temporal_unit_ttqa_tail}Distribution of the expected answers by temporal unit of the answer for the tail split of the TTQA dataset. Answers were transformed into numeric form. In the case of dates, they were converted into timestamps.}
\end{figure*}

\begin{figure*}
	\input{figures/answer_kde_per_temporal_unit_tot_arithmetic_None.pgf}
	\caption {\label{fig:answer_kde_per_temporal_unit_tot_arithmetic}Distribution of the expected answers by temporal unit of the answer for the arithmetic split of the ToT dataset. Answers were transformed into numeric form. In the case of dates, they were converted into timestamps.}
\end{figure*}

\begin{figure*}
	\input{figures/answer_kde_per_temporal_unit_tot_arithmetic_cluster_label.pgf}
	\caption {\label{fig:answer_kde_per_temporal_unit_tot_arithmetic_clustered}Distribution of the expected answers by temporal unit of the answer for the arithmetic split of the ToT dataset. If answers were clustered, clusters are highlighted by different colours. Answers were transformed into numeric form. In the case of dates, they were converted into timestamps.}
\end{figure*}

\section{BERTScore}
\label{sec:bertscore}

We did not consider similarity-based metrics, as they tend to return high similarity scores for digits, regardless of how close they are to each other, as shown in Tab.~\ref{tab:bertscore}.

\begin{table}
	\centering
	\begin{tabular}{@{}rrr@{}}
		\toprule
		Expected & Predicted & BERTScore \\
		\midrule
		1        & 1         & 1.0000    \\
		1        & 2         & 0.9998    \\
		1        & 10        & 0.9992    \\
		1        & 100       & 0.9987    \\
		\bottomrule
	\end{tabular}
	\caption{BERTScore for some predictions. Scores were rounded to the last four digits.}
	\label{tab:bertscore}
\end{table}

\section{Results extended}
\label{sec:appendix-results}
Results in tabular form are listed in Tab.~\ref{tab:tot-results-appendix} for ToT and in Tab.~\ref{tab:ttqa-results-appendix} for TTQA. Both tables include baseline experiments predicting the mean and the median of the respective data's split. 

We can see that for TTQA, the mean baseline's sMAPE is relatively high and underperforms all models. Only \Llama with zero-shot prompting underperforms the median baseline in the tail split. sMAPE shows that it favours small errors and that a simple baseline performs much worse than LLMs. The median baseline, according to MASE, outperforms two models in the head split. In the test split, 7 out of 12 models are worse than the baseline in the tail split, which is interesting given that the tail split contains less popular topics on Wikipedia for which it is harder to have good temporal domain knowledge. On the other hand, seeing that the median rather than the mean baseline ranks high should make us careful whether outliers skew model performance.

We see a similar pattern in the ToT dataset. According to sMAPE, only three models are slightly worse than the mean and median baseline when being evaluated on the arithmetic split. Regarding the semantic split, all models outperform both baselines according to sMAPE. According to MASE, both baselines are the worst performing on the arithmetic split. Regarding the semantic split, only Llama-3.3-70B outperforms the mean baseline according to MASE. No model has a better MASE than the median baseline. Since the dataset is synthetic, the variety in the data is low. An inspection of the data reveals that expected answers gravitated around similar answers (198X). It puts into perspective why so many models have very low MASE. Tasks in the semantic split require the extraction of dates from a graph, and these dates exhibit minimal variance, making the median baseline surprisingly effective, which explains the models' good performance otherwise.

Fig.~\ref{fig:smape_em_scatter} is a scatter plot with the results comparing EM and sMAPE, and Fig.~\ref{fig:mase_em_scatter} is a scatter plot comparing EM and MASE.

EM is defined for all QA pairs; sMAPE and MASE are not. sMAPE is not defined for dates or times. Since it has a maximum value, namely 100\%, it is defined even if the answer of the model is not parsable. MASE does not have this property as it has no upper bound. Instead, it is defined for dates and times. Tab.~\ref{tab:tot-defined-errors} lists the number of QA pairs in the ToT dataset for which either metric is defined, and Tab.~\ref{tab:ttqa-defined-errors} does the same for the TTQA dataset.

\subsection{Arithmetic errors in reasoning}

\begin{figure*}
	\centering
	\resizebox{\linewidth}{!}{
    \input{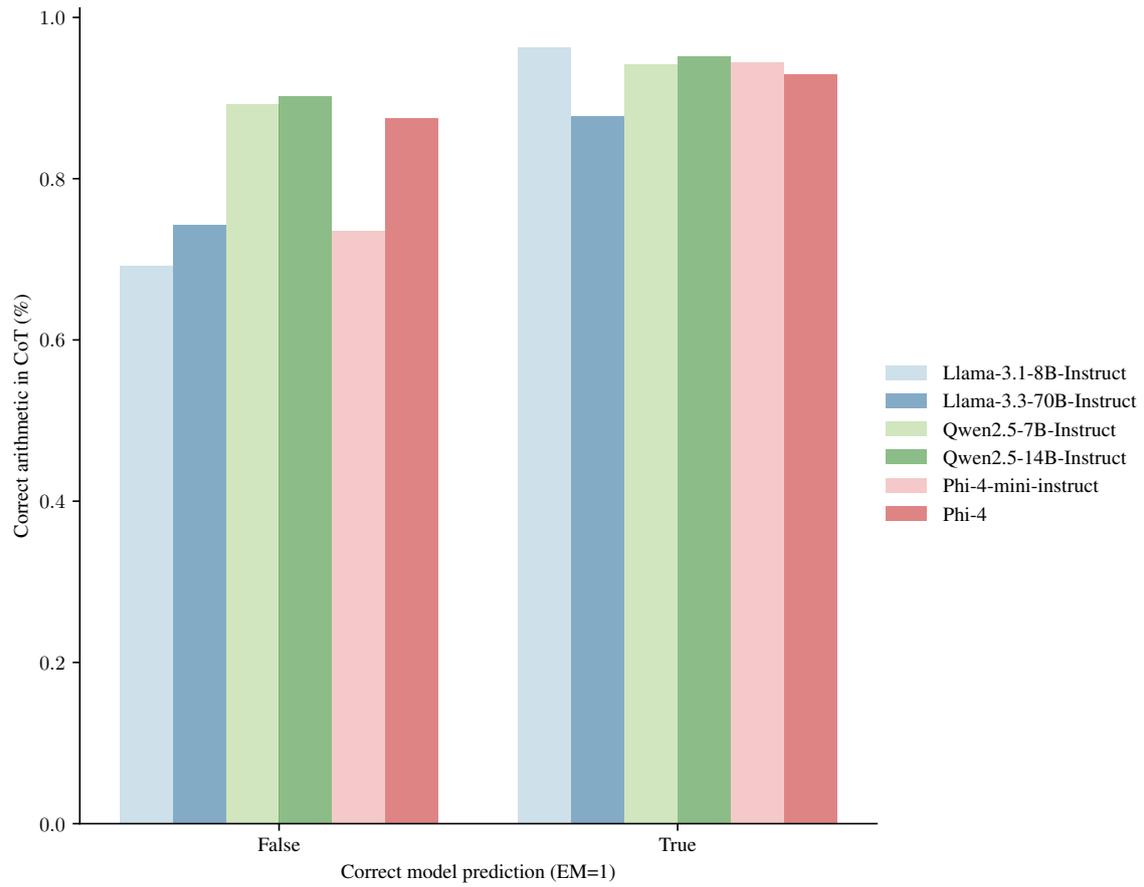}
    }
	\caption{\label{fig:cot-ari}Relationship between correct arithmetic in CoT-traces and correctness of final model response in the ToT dataset.}
\end{figure*}

ToT, especially the arithmetic split, requires strong mathematical reasoning skills. We investigated the relationship between correct arithmetic operations in the CoT-traces and a correct final model response. We used a regular expression to extract mathematical operations from the CoT-trace, calculated the left-hand side of each operation, and compared it with the correct answer on the right-hand side in the traces. The results are in Fig.~\ref{fig:cot-ari}.

A higher proportion of correct arithmetic calculations in the CoT-traces appears more often when the model's final response was correct. This relationship is strong in smaller models and in Llama models. This relationship is the smallest for Qwen models.

\begin{table*}
	\centering
	\begin{tabular}{@{}lllrrr@{}}
		\toprule
		Split                     & Model                             & Prompting & EM ($\uparrow$) & sMAPE ($\downarrow$) & MASE ($\downarrow$) \\
		\midrule
		\multirow[t]{12}{*}{head} & \multirow[t]{2}{*}{Llama-3.1-8B}  & few shot  & 75.34           & 17.02                & 0.50                \\
		                          &                                   & zero shot & 63.37           & 33.11                & 0.25                \\
		                          & \multirow[t]{2}{*}{Llama-3.3-70B} & few shot  & \bfseries 83.71 & 6.58                 & 0.20                \\
		                          &                                   & zero shot & 74.62           & 20.59                & \bfseries 0.17      \\
		                          & \multirow[t]{2}{*}{Phi-4-mini}    & few shot  & 77.05           & 7.03                 & 0.52                \\
		                          &                                   & zero shot & 73.09           & 12.65                & 7.45                \\
		                          & \multirow[t]{2}{*}{Phi-4}         & few shot  & 79.39           & 6.13                 & 0.34                \\
		                          &                                   & zero shot & 61.30           & 29.36                & 0.29                \\
		                          & \multirow[t]{2}{*}{Qwen2.5-7B}    & few shot  & 79.12           & 7.77                 & 4.47                \\
		                          &                                   & zero shot & 77.77           & 8.05                 & 0.91                \\
		                          & \multirow[t]{2}{*}{Qwen2.5-14B}   & few shot  & 81.73           & \bfseries 4.28       & 0.63                \\
		                          &                                   & zero shot & 76.33           & 9.21                 & 3.91                \\
                                  & Baseline, mean                    & -         & 0.00            & 37.52                & 7.35                \\
                                  & Baseline, median                  & -         & 0.00            & 34.24                & 2.25                \\
		\multirow[t]{12}{*}{tail} & \multirow[t]{2}{*}{Llama-3.1-8B}  & few shot  & 69.10           & 18.52                & 1.17                \\
		                          &                                   & zero shot & 57.14           & 36.29                & 1.66                \\
		                          & \multirow[t]{2}{*}{Llama-3.3-70B} & few shot  & 79.19           & 7.50                 & 0.19                \\
		                          &                                   & zero shot & 66.30           & 22.51                & 8.72                \\
		                          & \multirow[t]{2}{*}{Phi-4-mini}    & few shot  & 70.81           & 8.89                 & 8.87                \\
		                          &                                   & zero shot & 70.34           & 14.04                & 27.06               \\
		                          & \multirow[t]{2}{*}{Phi-4}         & few shot  & 77.48           & 7.96                 & 10.53               \\
		                          &                                   & zero shot & 61.18           & 29.58                & 11.41               \\
		                          & \multirow[t]{2}{*}{Qwen2.5-7B}    & few shot  & 73.29           & 8.20                 & 2.03                \\
		                          &                                   & zero shot & 70.81           & 10.62                & 23.05               \\
		                          & \multirow[t]{2}{*}{Qwen2.5-14B}   & few shot  & \bfseries 80.12 & \bfseries 4.89       & \bfseries 0.17      \\
		                          &                                   & zero shot & 73.60           & 8.60                 & 16.09               \\
                                  & Baseline, mean                    & -         & 0.00            & 38.67                & 123.88              \\
                                  & Baseline, median                  & -         & 0.00            & 31.21                & 2.29                \\
		\bottomrule
	\end{tabular}

	\caption{Model performance on the TTQA subset. The best performance per metric and split is bold.}
	\label{tab:ttqa-results-appendix}
\end{table*}

\begin{table*}
	\centering
	\begin{tabular}{@{}lllrrr@{}}
		\toprule
		Split                           & Model                             & Prompting & EM ($\uparrow$) & sMAPE ($\downarrow$) & MASE ($\downarrow$) \\
		\midrule
		\multirow[t]{12}{*}{arithmetic} & \multirow[t]{2}{*}{Llama-3.1-8B}  & few shot  & 20.57           & 23.66                & 2.23                \\
		                                &                                   & zero shot & 14.47           & 35.80                & 2.03                \\
		                                & \multirow[t]{2}{*}{Llama-3.3-70B} & few shot  & 49.70           & 10.54                & \bfseries 0.15      \\
		                                &                                   & zero shot & 41.34           & 13.53                & 0.37                \\
		                                & \multirow[t]{2}{*}{Phi-4-mini}    & few shot  & 26.38           & 21.19                & 1.00                \\
		                                &                                   & zero shot & 19.59           & 38.37                & 3.47                \\
		                                & \multirow[t]{2}{*}{Phi-4}         & few shot  & 55.71           & \bfseries 7.12       & 0.17                \\
		                                &                                   & zero shot & \bfseries 58.56 & 9.03                 & 0.16                \\
		                                & \multirow[t]{2}{*}{Qwen2.5-7B}    & few shot  & 31.00           & 20.85                & 1.06                \\
		                                &                                   & zero shot & 25.30           & 34.11                & 0.46                \\
		                                & \multirow[t]{2}{*}{Qwen2.5-14B}   & few shot  & 43.21           & 9.21                 & 0.71                \\
		                                &                                   & zero shot & 44.00           & 10.88                & 0.52                \\
                                        & Baseline, mean                    & -         & 0.00            & 24.26                & 4.60                \\
                                        & Baseline, median                  & -         & 0.00            & 23.14                & 2.96                \\
		\multirow[t]{12}{*}{semantic}   & \multirow[t]{2}{*}{Llama-3.1-8B}  & few shot  & 72.69           & 10.43                & 1.18                \\
		                                &                                   & zero shot & 67.11           & 14.19                & 11.34               \\
		                                & \multirow[t]{2}{*}{Llama-3.3-70B} & few shot  & 89.43           & 4.61                 & \bfseries 0.11      \\
		                                &                                   & zero shot & \bfseries 90.16 & 6.09                 & 0.19                \\
		                                & \multirow[t]{2}{*}{Phi-4-mini}    & few shot  & 65.64           & 10.54                & 1.04                \\
		                                &                                   & zero shot & 56.68           & 17.64                & 1.27                \\
		                                & \multirow[t]{2}{*}{Phi-4}         & few shot  & 83.55           & \bfseries 4.01       & 0.32                \\
		                                &                                   & zero shot & 80.32           & 6.39                 & 0.46                \\
		                                & \multirow[t]{2}{*}{Qwen2.5-7B}    & few shot  & 63.00           & 8.47                 & 1.19                \\
		                                &                                   & zero shot & 59.32           & 11.57                & 1.27                \\
		                                & \multirow[t]{2}{*}{Qwen2.5-14B}   & few shot  & 72.98           & 8.79                 & 0.50                \\
		                                &                                   & zero shot & 67.99           & 11.72                & 0.61                \\
                                        & Baseline, mean                    & -         & 0.00            & 18.14                & 0.26                \\
                                        & Baseline, median                  & -         & 0.10            & 18.07                & \bfseries *0.04     \\
		\bottomrule
	\end{tabular}
	\caption{Model performance on the ToT subset. The best performance of a model per metric and split is bold. A baseline exceeding a model is made bold with an additional asterisk.}
	\label{tab:tot-results-appendix}
\end{table*}

\begin{table*}
	\centering
	\begin{tabular}{@{}llrrr@{}}
		\toprule
		                                  &           & \multicolumn{3}{c}{\# of defined errors}                \\
		\cmidrule(lr){3-5}
		Model                             & Prompting & EM                                       & sMAPE & MASE \\
		\midrule
		\multirow[t]{2}{*}{Llama-3.1-8B}  & few shot  & 1737                                     & 1373  & 1530 \\
		                                  & zero shot & 1737                                     & 1373  & 1225 \\

		\multirow[t]{2}{*}{Llama-3.3-70B} & few shot  & 1737                                     & 1373  & 1667 \\
		                                  & zero shot & 1737                                     & 1373  & 1417 \\

		\multirow[t]{2}{*}{Phi-4-mini}    & few shot  & 1737                                     & 1373  & 1722 \\
		                                  & zero shot & 1737                                     & 1373  & 1668 \\

		\multirow[t]{2}{*}{Phi-4}         & few shot  & 1737                                     & 1373  & 1680 \\
		                                  & zero shot & 1737                                     & 1373  & 1341 \\

		\multirow[t]{2}{*}{Qwen2.5-7B}    & few shot  & 1737                                     & 1373  & 1706 \\
		                                  & zero shot & 1737                                     & 1373  & 1708 \\

		\multirow[t]{2}{*}{Qwen2.5-14B}   & few shot  & 1737                                     & 1373  & 1727 \\
		                                  & zero shot & 1737                                     & 1373  & 1700 \\
		\bottomrule
	\end{tabular}
	\caption{Number of QA pairs of the TTQA dataset for which each metric is defined. EM is defined for each question. sMAPE is not defined for dates and is set to 100\% if errors are not parsable. MASE is defined for all questions, but is not defined if the model's answer is not parsable.}
	\label{tab:ttqa-defined-errors}
\end{table*}

\begin{table*}
	\centering
	\begin{tabular}{@{}llrrr@{}}
		\toprule
		                                  &           & \multicolumn{3}{c}{\# of defined errors}                \\
		\cmidrule(lr){3-5}
		Model                             & Prompting & EM                                       & sMAPE & MASE \\
		\midrule
		\multirow[t]{2}{*}{Llama-3.1-8B}  & few shot  & 1697                                     & 1369  & 1575 \\
		                                  & zero shot & 1697                                     & 1369  & 1527 \\
		\multirow[t]{2}{*}{Llama-3.3-70B} & few shot  & 1697                                     & 1369  & 1581 \\
		                                  & zero shot & 1697                                     & 1369  & 1524 \\
		\multirow[t]{2}{*}{Phi-4-mini}    & few shot  & 1697                                     & 1369  & 1618 \\
		                                  & zero shot & 1697                                     & 1369  & 1438 \\
		\multirow[t]{2}{*}{Phi-4}         & few shot  & 1697                                     & 1369  & 1640 \\
		                                  & zero shot & 1697                                     & 1369  & 1595 \\
		\multirow[t]{2}{*}{Qwen2.5-7B}    & few shot  & 1697                                     & 1369  & 1600 \\
		                                  & zero shot & 1697                                     & 1369  & 1499 \\
		\multirow[t]{2}{*}{Qwen2.5-14B}   & few shot  & 1697                                     & 1369  & 1660 \\
		                                  & zero shot & 1697                                     & 1369  & 1648 \\
		\bottomrule
	\end{tabular}
	\caption{Number of QA pairs of the ToT dataset for which each metric is defined. EM is defined for each question. sMAPE is not defined for dates and is set to 100\% if errors are not parsable. MASE is defined for all questions, but is not defined if the model's answer is not parsable.}
	\label{tab:tot-defined-errors}

\end{table*}

\begin{figure*}
	\centering
	\input{figures/smape_em_results.pgf}
	\caption{\label{fig:smape_em_scatter}Comparison of performance measured by sMAPE and EM.}
\end{figure*}

\begin{figure*}
	\centering
	\input{figures/mase_em_results.pgf}
	\caption{\label{fig:mase_em_scatter}Comparison of performance measured by MASE and EM.}
\end{figure*}

\section{Model responses}
\label{sec:model-responses}

We hypothesise that synthetic data training contributes to lower MASE scores. Here are two examples from ToT and TTQA to illustrate high MASE responses from \Qwen and \PhiModel.

\begin{promptbox}
	\textbf{TTQA question:} How old was Arntzen when she began to play for her first senior club?

	\textbf{\Qwen:}  6

	\textbf{Expected:} 16
\end{promptbox}

\Qwen needs to answer a question about the Norwegian handball player Emilie Arntzen. It should be clear that a professional handball player cannot be six years old.

\begin{promptbox}
	\textbf{ToT question:} It takes Sophia an average of 16 minutes and 46 seconds to bake 2 cakes. If she wants to bake 15 cakes at the same rate, it will take her X hours, Y minutes, and Z seconds. Report the values of X, Y and Z as a json of the form {"explanation": [object Object], "X": X, "Y": Y, "Z": Z}.

	\textbf{\PhiModel:}  {'X': 127, 'Y': 30, 'Z': 0}

	\textbf{Expected:} {'X': 2.0, 'Y': 5.0, 'Z': 45.0}
\end{promptbox}

By roughly averaging, we can see that \~17 minutes × 15 must be below 300 minutes (20 × 15). Instead, \PhiModel estimates that it must take more than 5 days.

\end{document}